\documentclass[letterpaper, 10 pt, conference]{ieeeconf}  

\IEEEoverridecommandlockouts                              

\usepackage{graphics} 
\usepackage{epsfig} 
\usepackage{mathptmx} 
\usepackage{times} 
\usepackage{amsmath} 
\usepackage{amssymb}  

\usepackage{graphicx}
\usepackage{multirow}
\usepackage{booktabs}
\usepackage{xcolor}
\usepackage{eucal}

\title{\LARGE \bf
UAV-Sim: NeRF-based Synthetic Data Generation for UAV-based Perception}

\author{Christopher Maxey$^{*}$\:\textsuperscript{\rm 1,2}, Jaehoon Choi$^{*}$\:\textsuperscript{\rm 2}, Hyungtae Lee$^{1}$, Dinesh Manocha$^{2}$, and Heesung Kwon$^{1}$\\
$^{1}$DEVCOM Army Research Laboratory $^{2}$University of Maryland
\thanks{* These two authors contributed equally}%
\thanks{Correspondence to cmaxey@umd.edu, kevchoi@umd.edu}%
}

\begin{document}

\maketitle
\thispagestyle{empty}
\pagestyle{empty}

\begin{abstract}


Tremendous variations coupled with large degrees of freedom in UAV-based imaging conditions lead to a significant lack of data in adequately learning UAV-based perception models. Using various synthetic renderers in conjunction with perception models is prevalent to create synthetic data to augment the learning in the ground-based imaging domain. However, severe challenges in the austere UAV-based domain require distinctive solutions to image synthesis for data augmentation. In this work, we leverage recent advancements in neural rendering to improve static and dynamic novel-view UAV-based image synthesis, especially from high altitudes, capturing salient scene attributes. Finally, we demonstrate a considerable performance boost is achieved when a state-of-the-art detection model is optimized primarily on hybrid sets of real and synthetic data instead of the real or synthetic data separately. 

\end{abstract}

\section{INTRODUCTION}

Unmanned aerial vehicle (UAV)-based perception, such as recognizing objects of interest in real-time, is a core problem in numerous applications spanning various civil and military fields. Modern UAVs can be equipped with low power, mobile GPUs that can run efficient yet state-of-the-art (SOTA) neural networks, for instance, YOLOv8 \cite{yolov8}. Utilizing the compact networks embedded on small UAVs enables important tasks, such as search and rescue for disaster relief \cite{disaster}, highlighting points of interest, such as people or vehicles in surveillance and reconnaissance \cite{surveillance}, and assessing the conditions or recognizing an occurrence of specific activities, or events in a certain region of interest \cite{traffic,Agriculture}.  


Due to the large degrees of freedom of the UAV-based imaging domain,  procuring sufficient training data for UAV-based perception can be an onerous task. Various challenges, including flight restrictions, security and safety issues, and weather conditions, further compound the complications of UAV-based data collection. In particular, setting up adequate backgrounds or surroundings so that the scene can closely reflect a real-world scenario can be costly and time-consuming. Compounded by the data-hungry nature of UAV-based perception algorithms, acquiring diverse training data representing real-world scenes and backgrounds becomes a significant barrier against learning efficient UAV-based learning models. 

\begin{figure}[t]
\centering
\includegraphics[trim= 5mm 5mm 5mm 5mm, clip, width=\linewidth]{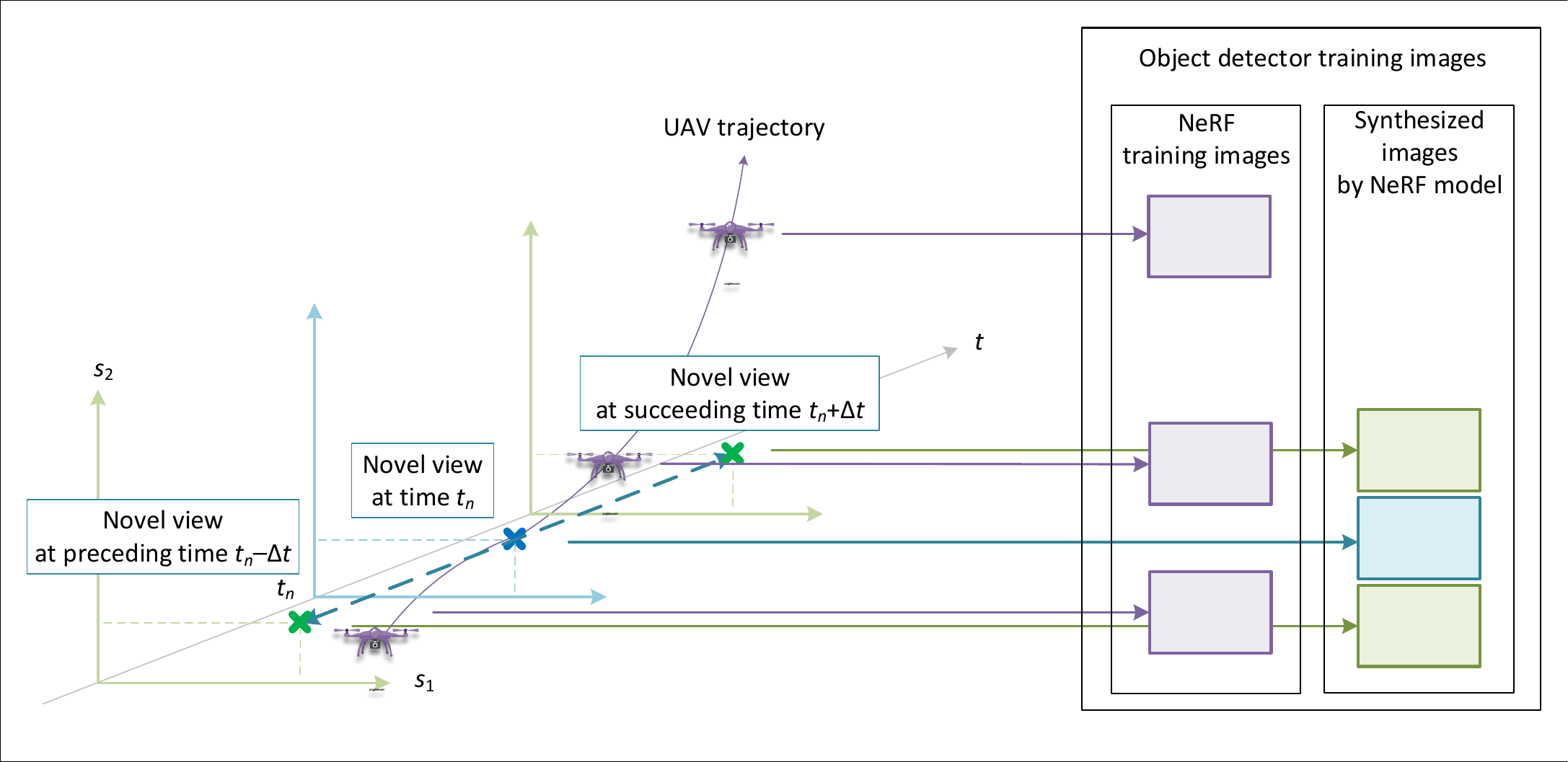}
\caption{{\bf Constructing a training set of UAV-Sim.} On the left, camera poses are shown in spatial ($s_\text{1}$, $s_\text{2}$) and temporal ($t$) coordinates. Images captured with the corresponding camera poses are shown on the right. UAV-Sim generates synthetic images (blue x mark) at spatial locations on the UAV trajectory using NeRF models trained on the limited original UAV-based images. To cope with the dynamic nature of UAV-based images, UAV-Sim also generates synthetic images (green x marks) at different times at a certain location on the trajectory. The training set for an object detector can be constructed with a hybrid set of original real images and novel-view synthetic images.}
\label{concept_figure}
\vspace{-3mm}
\end{figure}


To help alleviate some of the burdens of learning data-hungry algorithms in the UAV-based domain, creating synthetic data has become an active area of research.  A straightforward approach involves generating high-quality synthetic data using synthetic renderers \cite{Unreal4,Simulated_photoreal} or generative models (GANs \cite{GAN}, or diffusion \cite{stable-diffusion}).  However, they still suffer from an issue known as \emph{domain gap} \cite{domain_gap} in which synthetic data in training and the test data differ in fine details of the scene, including appearance, texture, etc., resulting in performance decrease during testing. Recent advancements in scene synthesis, such as neural radiance fields (NeRFs) \cite{NeRF}, have allowed for high-fidelity scene reconstruction \cite{NeRF++} and novel-view image synthesis \cite{Instant-NGP,Mip-nerf360}, given a limited set of training images from the target scene.  NeRFs can create high-quality novel-view imagery with enhanced realism that closely matches the target scene, effectively augmenting training data and thus enhancing model learning. 



\noindent\textbf{Main Results:} Our main goal is to learn NeRF-based models in the UAV-based imaging domain that can create novel-view images from previously unseen camera positions capturing salient attributes of the scene. As shown in Fig. 1, our approach is designed to augment datasets by generating synthetic images that are applicable to both static and dynamic scenes relevant to the object detection, as well as action and activity recognition. In this work, we show that using NeRFs is particularly beneficial in supplementing UAV-based data as most UAV-based benchmarks do not sufficiently represent huge variations in the UAV-based imaging domain.


We also explore the challenges of reconstructing novel-view dynamic scenes with fidelity on par with original scenes by enhancing an existing dynamic NeRF algorithm \cite{K-planes}. Our approach is able to better capture the details of dynamic scenes from the UAV-based data for human action recognition called the Okutama-Action dataset \cite{Okutama}. In UAV-based data, it is not possible to use data-driven depth, optical flow, and motion segmentation masks which are typically used by previous dynamic NeRF methods \cite{Dynibar,Nerfies,Dnerf}.  
To the best of our knowledge, our work is one of the first attempts to augment UAV-based data captured with challenging conditions including relatively high altitudes and far range, steep view angles, and unstable imaging conditions. Some novel components of our work include:  


\begin{itemize}


\item Development of a model optimization pipeline for UAV-based perception using NeRF integrating unseen critical salient attributes into model fine-tuning with self-generated bounding box annotation.
\item Extension of the optimization pipeline to dynamic UAV-based scene using extended dynamic NeRF to enhance recognition of objects in motion.
\item Demonstration of the overall benefit of leveraging NeRF for data augmentation in the UAV-based perception, showing a 55.85\% improvement in mAP for static scenes and a 12.4\% enhancement for dynamic scenes. 
\end{itemize}

\section{Related Work}
The most prevalent approach for generating high-quality synthetic data involves the use of simulators such as game engine.
Most simulators \cite{Unreal4,Airsim,Simulated_photoreal} are capable of synthesizing high-quality data for training deep learning-based models. Nevertheless, most of them exhibit two limitations: a domain gap \cite{domain_gap} and the need for skilled human graphics experts. 
Recently, neural rendering methods that solely rely on image data and camera poses have demonstrated remarkable levels of photo-realistic rendering quality. In particular, NeRF \cite{NeRF} introduce the concept of the neural radiance field representation and apply differentiable volume rendering. Various methods have emerged for improving training efficiency \cite{Instant-NGP} and visual quality \cite{Mip-nerf360}. Some works \cite{Mega-NeRF,Drone-NeRF} primarily focus on enhancing NeRF for large-scale drone footage, emphasizing improvements at training efficiency and rendering quality. Our main focus, however, is on applying NeRF to generate synthetic data and to improve perception algorithms for UAV images.

As can be pertinent to UAV footage, NeRF research also encompasses rendering dynamic scenes \cite{WildNerf}.  One common approach is to model a static canonical scene and a deformation network to adjust time dependent input rays to match the canonical scene \cite{Nerfies,Dnerf}.  This effectively guides the NeRF on how the scene should look at a given time with respect to a pose at a fixed time.  DynIBaR \cite{Dynibar} utilizes features extracted from the images and stacks said features along epipolar lines for static portions of the scene and along a learned deformable line for transient portions of the scene.  Another approach relies on explicitly representing a scene via data structures that store feature vectors based on locations within a scene.  This can be done with a full dimensional volume representation of a scene or by factoring a scene's volume into planes, both static and dynamic \cite{Hexplane,K-planes}.  Time-aware voxels in \cite{FastD} utilize a full 3D volume to represent the scene but also includes a deformation network for rays projected into the volume.  All of these methods focus on dynamic scenes that are from a ground perspective and typically have transient motions that are near to the camera. Dynamic imagery from distant cameras such as UAVs presents additional challenges.

\section{Our Method}

Our goal is to develop a new data augmentation method to improve performance for UAV-based object detection. In this paper, we propose using two Neural Radiance Fields (NeRFs) for either static or dynamic scenes can be incorporated into augmentation method. One NeRF is designed for static scenes, which are captured by UAVs and exclusively comprise static objects. The other is tailored for dynamic scenes that include moving people or pedestrians (Section~\ref{ssec:NeRFs}). To properly leverage generated images by NeRF models, we also designed a selection strategy of camera poses and time for novel-view images (Section~\ref{ssec:selection_strategy}).


\begin{figure*}[t]
\centering
\includegraphics[width=0.85\linewidth]{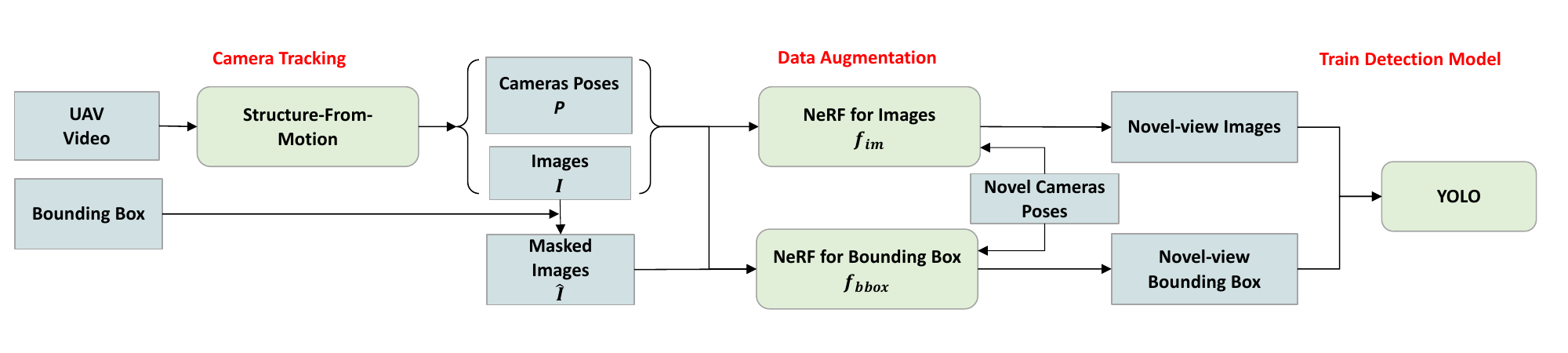}
\vspace{-7mm}
\caption{Overview of our training pipeline for static NeRF. Our method requires two NeRF models ($f_{im}$ and $f_{bbox}$) trained using either original images or masked images. When provided with novel camera poses, $f_{im}$ generates novel augmented images, while $f_{bbox}$ synthesizes novel masked images for the purpose of bounding box extraction. Then, we can acquire novel-view images along with their corresponding bounding boxes to train the object detector.}
\label{Static_NeRF_figure}
\vspace{-5mm}
\end{figure*}

\begin{figure}[t]
\centering
\includegraphics[width=0.95\linewidth]{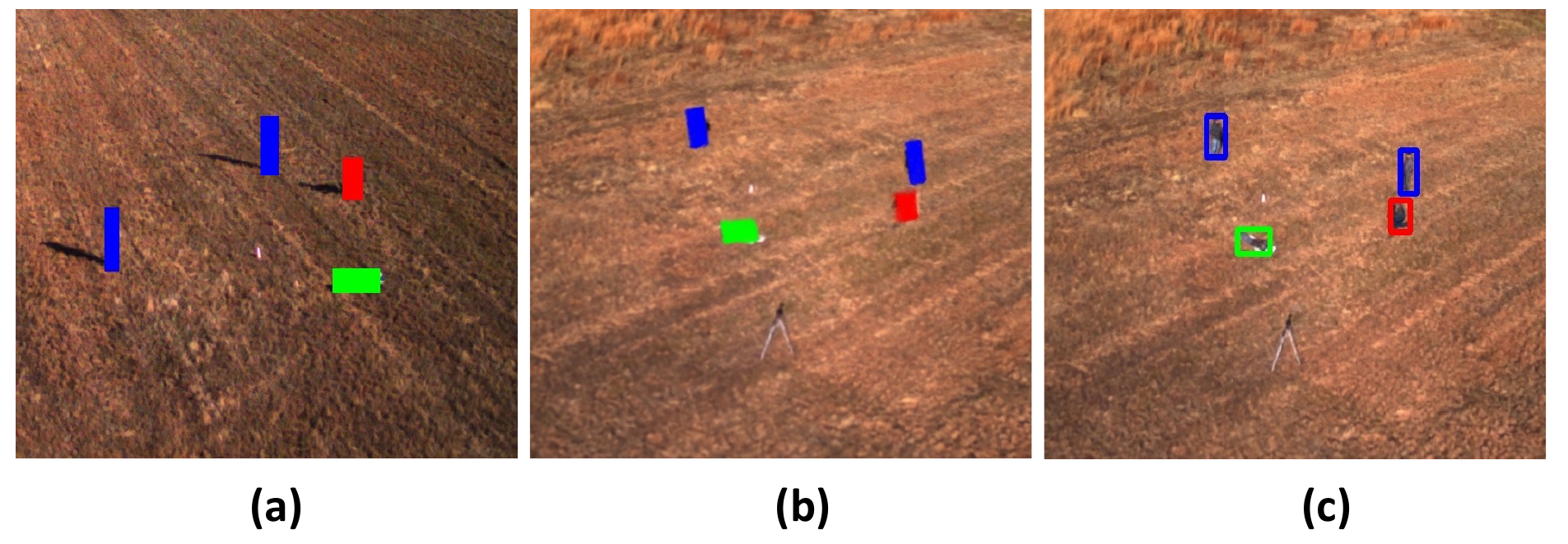}
\vspace{-2mm}
\caption{Examples of intermediate results of data augmentation pipeline. (a) Masked images. (b) Novel-view images generated by NeRF with bounding box masks. (c) Novel-view images with corresponding bounding boxes.  
}
\label{Static_NeRF_figure2}
\vspace{-6mm}
\end{figure}

\subsection{Novel-View Image Generation}
\label{ssec:NeRFs}


In this task, we address both static and dynamic scene synthesis, given the inherent characteristics of UAV-based imaging in real-world scenarios. A UAV captures the scene while in flight with objects of interest on the ground, static or in motion. Therefore, general NeRFs using images taken from multiple perspectives at the same time require special consideration to deal with dynamic scenes such as our scenario. In this subsection, we first describe NeRF-based data augmentation for object detection in static scenes and then cover alternative methods used for dynamic scenes.

\textbf{1) Static Scene Synthesis:} To generate a novel-view image, we employ a NeRF model, $f_\text{im}$, which is trained using original input images $I$ and their corresponding camera pose $P$, coupled with classical volume rendering techniques as described in \cite{Rendering_equation}. In our approach, the camera pose $P$ for a new-view image is calculated by Structure-from-motion (SfM)~\cite{pix-sfm}. Then, the novel-view images must be annotated by locating bounding boxes on objects of interest before training an object detector.

To estimate the bounding boxes in the novel-view images, we use a separate NeRF model, $f_\text{bbox}$. $f_\text{bbox}$ is trained to synthesize masked images corresponding to the object-of-interest regions in the original images from a novel viewpoint. The bounding box in the generated images can be acquired from the masked images without requiring human-in-the-loop annotation. For example, Fig. \ref{Static_NeRF_figure2}-(b) is the synthesized image from $f_\text{bbox}$ with bounding box masks and we can obtain the bounding boxes for each color blob as shown in Fig. \ref{Static_NeRF_figure2}-(c). Although it is possible to transform bounding boxes using depth maps and camera parameters from the training images to align them with novel-view images, the accuracy of the depth maps generated by NeRF is insufficient to achieve perfect alignment with small human objects, due to their volumetric rendering properties \cite{NeRF++}.

Following the generation of novel-view images along with their respective bounding boxes, we use this synthetic data for fine-tuning the state-of-the-art object detection model, \textit{e.g.}, YOLOv8 \cite{yolov8}. Our overall pipeline is shown in Fig.~\ref{Static_NeRF_figure}.For the NeRF architecture, we adopt the Nerfacto model from Nerfstudio \cite{nerfstudio} that combines various features from recent papers such as scene contraction \cite{Mip-nerf360}, hash grid encoding \cite{Instant-NGP}, and proposal network sampler \cite{Mip-nerf360}. Both NeRF models are represented by Multi-layer Pereceptron (MLP) networks, which takes a 3D point and viewing direction (or camera pose) as input and produce color and density as output.

\textbf{2) Dynamic Scene Synthesis: Considering Temporal Factor~} To cope with dynamic scenes, we employ the K-Planes dynamic NeRF algorithm \cite{K-planes}. This algorithm represents a scene explicitly by factoring the spatial and temporal dimensions of a 4D volume into 2D planes in a certain space and time. Specifically, the K-planes method uses six planes to represent dynamic scenes, where three planes are for the pairwise combinations of spatial dimensions and the rest are for each spatial dimension plus time, \textit{i.e.}, $xy$, $xz$, $yz$, $xt$, $yt$, and $zt$. When rendering the novel view image at a certain point in time, a 4D point (three spatial dimensions plus time) is projected onto each plane, and feature vectors from each plane are interpolated and multiplied to get a final vector for a given 4D location, as follows:


\begin{equation} \label{eq:1}
    f(\mathbf{q}) = \prod_{c \in C}f(\mathbf{q})_{c}
\end{equation}

\noindent wherein $c\in C$ is each plane in the set of all planes, $\mathbf{q}$ represents a 4D coordinate $(i,j,k,\tau)$, $f(\mathbf{q})_{c}$ is the specific feature vector for a given plane based on the projection of $\mathbf{q}$ and subsequent interpolation between plane cells, and $f(\mathbf{q})$ is the final feature vector passed to the NeRF models to output a location density and color.  The factored planes reduce the memory requirements for representing a scene while maintaining equivalent performance to full 4D representations.  K-Planes is among the state of the art for performance on datasets such as DNerf \cite{Dnerf} and DyNeRF \cite{Dynerf} and is chosen for its compact representation.


\textbf{3) Dynamic Scene Synthesis: Extended K-Planes~} To rectify the performance issues of stock K-Planes, we extend the algorithm in several ways. We empirically find out that factoring the 4D volume into a set of static spatial planes, $C_{s}$, and a set of dynamic spatial and temporal planes, $C_{d}$, helps to alleviate some of the issues with noisy camera poses and better separates the static and temporal elements of the scene.  With this new factorization, we extract feature vectors as before, multiplying the three static plane feature vectors together and separately multiplying the six dynamic plane feature vectors together, as follows:

\begin{equation} \label{eq:2}
f_{s}(\mathbf{q}) = \prod_{c \in C_{s}}f_{s}(\mathbf{q})_{c}, \;\;\;\;f_{d}(\mathbf{q}) = \prod_{c \in C_{d}}f_{d}(\mathbf{q})_{c}.
\end{equation}

Here, each temporal plane stores feature vectors of dimension $D + 1$, where the extra element is used for mask generation corresponding to the ground truth bounding boxes.  $f_{s}$ and $f_{d}$ are passed through an MLP along with the learned mask values from the three temporal planes.  The output of the MLP is a feature vector $f$ of dimension $D$.  This vector is passed as input to the decoder MLPs to output a density and color at $\mathbf{q}$, the same as in the stock algorithm.



We also introduced a cosine-similarity loss between the plane feature vectors for each location.  The loss is applied between each pair of static and dynamic spatial planes, as follows:

\begin{equation} \label{eq:3}
    \CMcal{L}_\text{cos} = \sum_{c\in C_{spatial}}|d_\text{cos}(f_{s}(\mathbf{q})_c,~f_{d}(\mathbf{q})_c)|,
\end{equation}

\noindent where $d_\text{cos}(\cdot,\cdot)$ represents cosine similarity and $C_{spatial}$ are the spatial dimension combination of planes for both static and dynamic planes.  The absolute value of the cosine similarity is minimized in order to separate the learned features for the static and dynamic planes.  This loss helps to reduce cross learning of static features in the dynamic planes due to pose inaccuracies being interpreted as temporal movement between frames.

Finally, we take inspiration from the mask loss in \cite{Dynibar} to implement a similar function.  Ground truth bounding boxes are used as a segmentation mask for the dynamic regions of a scene.  Given the separation of static and dynamic planes as mentioned above, we can apply this mask to the back propagation of the gradients for each set of planes such that only gradients from static pixels influence the static planes and only gradients from dynamic pixels influence the dynamic planes.  


\subsection{Novel Camera Pose Selection}
\label{ssec:selection_strategy}


To generate novel-view images, it is required to identify the camera pose $P$ that defines the view. For static scens, we randomly sample new camera poses based on input poses for the NeRF. In order to diversify our dataset and ensure robust training, we incorporate various poses that encompass factors like altitude, camera viewing angle, and radius rotation of the camera circle. Subsequently, we interpolate these novel trajectories to render the novel-view images. 

When determining new camera poses for dynamic scenes, the novel pose cannot deviate significantly from the input poses to NeRF. Considering this limitation, we randomly selected $N$ spatial locations on the camera trajectory of the training images. For each spatial location, we generated images at three different times for dynamic scenes. The three different times represent 1) the time when the camera is at the location, 2) the preceding time by $\Delta\:t$, and 3) the succeeding time by $\Delta\:t$ (\textit{i.e.}, $t$, $t\pm\Delta\:t$). $\Delta\:t$ is defined as the recipocal of a frame rate of UAV-based footage. As a result, using the selected camera poses and the trained NeRF models, we can generate $N$ novel-view images for static scenes and 3$N$ novel-view images for dynamic scenes.

\section{Implementation and Performance}

\subsection{Datasets} 
We conducted experiments using two datasets, the Archangel and Okutama-Action datasets, both of which entail detecting peoplee in various poses standing in open areas. The Archangel dataset \cite{Archangel} is chosen for static scenes as it represents a fairly simple scenario of people maintaining poses in a field.  This offers a strong baseline as both the background and target objects lack complicating details.  The Okutama-Action dataset \cite{Okutama} is chosen for dynamic scenes as it represents a more realistic scenario for UAV drone footage.  Background details are more complicated than Archangel and the moving people in the scene reflect the reality of most situations in which not every object is static.
\subsubsection{Archangel Dataset}

The Archangel dataset \cite{Archangel} is a unique hybrid dataset captured using  Unmanned Aerial Vehicles (UAVs), under similar imaging conditions in both real and synthetic domains. It provides metadata detailing camera positions, including UAV altitudes and the radii of rotation circles, for each individual image. We only use the Archangel-Real sub-dataset for our experiment. Archangel-Real, collected from a real-world environment, includes a group of real human subjects as targets, with each individual assuming one of three potential poses: standing, kneeling, and prone. Both the altitude and radius of the rotation circle were varied within the range of 15 to 50 meters, with 5-meter increments.

\subsubsection{Okutama-Action Dataset}

The Okutama-Action dataset \cite{Okutama} includes video from a UAV within a baseball stadium with multiple human ``agents" in the scene performing a variety of single and multi-agent actions.  There are up to 9 actors per scene and 12 different action categories on display.  The drones range from 10 to 45 meters in altitude with a camera angle set to either 45 or 90 degrees.  The dataset consists of two separate set of scenes, each one of which with its own unique pilot for the drone.  Videos are captured at both morning and noon time frames.  In light of all of this, Okutama-Action presents a challenging dataset for a variety of tasks, including scene reconstruction and novel view synthesis with NeRF algorithms.

\begin{table*}[t]
    \begin{center}
    \caption{\label{table_static_nerf} AP Comparison on the different altitudes of YOLOv8 family}
    \resizebox{0.95\textwidth}{!}{
    \begin{tabular}{l|c|c|cccc|cccc|cccc|cccc}
    \toprule
    \multirow{2}{*}{Data}& \multirow{2}{*}{Model} & \multirow{2}{*}{Metric} & \multicolumn{4}{c|}{Altitude 35m} & \multicolumn{4}{c|}{Altitude 40m} & \multicolumn{4}{c|}{Altitude 45m} & \multicolumn{4}{c}{Altitude 50m} \\
    & & &  Standing & Kneeling  & Prone & All & Standing & Kneeling & Prone & All& Standing & Kneeling & Prone & All & Standing & Kneeling & Prone & All \\ \hline
    Real & YOLOv8n & $\text{mAP}_{50}$ & 0.877 & 0.815 & 0.937 & 0.876 & 0.756 & 0.674 & 0.914 & 0.782 & 0.663 & 0.602 & 0.893 & 0.719 & 0.491 & 0.351 & 0.748 & 0.530 \\
    Synthetic & YOLOv8n & $\text{mAP}_{50}$ & \textbf{0.992} & 0.976 & 0.969 & 0.979 & 0.977 & 0.967 & 0.948 & 0.964 & 0.977 & 0.905 & 0.898 & 0.927 & 0.856 & 0.555 & 0.841 & 0.750\\
    R + S & YOLOv8n & $\text{mAP}_{50}$ & 0.978 & \textbf{0.983} & \textbf{0.993} & \textbf{0.985} & \textbf{0.980} & \textbf{0.984} & \textbf{0.975} & \textbf{0.980} & \textbf{0.977} & \textbf{0.957} & \textbf{0.944} & \textbf{0.959} & \textbf{0.864} & \textbf{0.670} & \textbf{0.942} & \textbf{0.826} \\
    \midrule
    Real & YOLOv8n & $\text{mAP}_{50:95}$ & 0.507 & 0.372 & \textbf{0.447} & 0.442 & 0.377 & \textbf{0.444} & 0.312 & 0.374 & 0.387 & 0.284 & 0.293 & 0.321 & 0.183 & 0.106 & 0.257 & 0.182  \\
    Synthetic & YOLOv8n & $\text{mAP}_{50:95}$  & 0.455 & \textbf{0.423} & 0.343 & 0.407 & 0.494 & 0.405 & 0.366 & 0.422 & 0.475 & 0.329 & 0.279 & 0.361 & 0.255 & 0.136 & 0.255 & 0.215 \\
    R + S & YOLOv8n & $\text{mAP}_{50:95}$  & \textbf{0.546} & 0.418 & 0.389 & \textbf{0.451} & \textbf{0.546} & 0.433 & \textbf{0.384} & \textbf{0.454} & \textbf{0.535} & \textbf{0.393} & \textbf{0.327} & \textbf{0.418} & \textbf{0.269} & \textbf{0.190} & \textbf{0.345} & \textbf{0.268} \\
    \bottomrule
    Real & YOLOv8s & $\text{mAP}_{50}$ & 0.879 & 0.817 & 0.988 & 0.895 & 0.832 & 0.763 & 0.983 & 0.859 & 0.767 & 0.707 & 0.959 & 0.811 & 0.586 & 0.490 & \textbf{0.940}  & 0.672 \\
    Synthetic & YOLOv8s & $\text{mAP}_{50}$ & \textbf{0.994} & \textbf{0.958} & \textbf{0.973} & 0.975 & \textbf{0.991} & \textbf{0.977} & \textbf{0.991} & \textbf{0.987} & \textbf{0.992} & \textbf{0.950} & 0.928 & \textbf{0.957} & \textbf{0.927} & \textbf{0.880} & 0.890 & \textbf{0.899} \\
    R + S & YOLOv8s & $\text{mAP}_{50}$ & 0.992 & 0.949 & 0.994 & \textbf{0.978} & 0.965 & 0.955 & 0.977 & 0.966 & 0.944 & 0.929 & \textbf{0.964} & 0.946 & 0.843 & 0.825 & 0.906 & 0.858 \\
    \midrule
    Real & YOLOv8s & $\text{mAP}_{50:95}$ & 0.537 & 0.367 & \textbf{0.471} & 0.459 & 0.497 & 0.315 & \textbf{0.429} & 0.414 & 0.454 & 0.328 & 0.368 & 0.383 & 0.229 & 0.161 & 0.330 & 0.240  \\
    Synthetic & YOLOv8s & $\text{mAP}_{50:95}$ & 0.551 & \textbf{0.464} & 0.391 & 0.468 & 0.549 & 0.428 & 0.390 & 0.455 & 0.578 & 0.403 & 0.306 & 0.429 & 0.311 & 0.253 & 0.310 & 0.291 \\
    R + S & YOLOv8s & $\text{mAP}_{50:95}$ & \textbf{0.589} & 0.434 & 0.452 & \textbf{0.492} & \textbf{0.578} & \textbf{0.459} & 0.421 & \textbf{0.486} & \textbf{0.589} & \textbf{0.416} & \textbf{0.383} & \textbf{0.463} & \textbf{0.330} & \textbf{0.263} & \textbf{0.365} & \textbf{0.319} \\
    \bottomrule
    \end{tabular}}
    \end{center}
    ``Real" denotes the detection performance achieved through training with Archangel-Real data collected at an altitude of 15m. ``Synthetic" represents the performance achieved through training with synthetic data generated by NeRF. ``R+S" indicates the detection accuracy attained through training with a combined dataset that includes both Archangel-Real and NeRF-generated synthetic data. Compared to YOLOv8n trained with ``Real" dataset, we improve the $\text{mAP}_{50}$ metric by 55.85 \% and the $\text{mAP}_{50:95}$ metric by 47.25 \% at 50m altitude.
    \vspace{-6mm}
\end{table*}

\subsection{Implementation Details}
For each experiment, we employ the Nerfacto and K-Planes within the Nerfstudio \cite{nerfstudio}.
The Nerfacto model are trained for 30K iterations, requiring up to half an hour on an NVIDIA RTX 3090 GPU, while K-planes models are trained for 100K iterations, taking up to 3.5 hours on the same GPU.
Hyperparameters for losses are kept as their default values.  The learning rate is lowered to 1e-3 for both network proposals and plane features.  Three plane sizes are used with dimensions $(x,y,z,t) = (128,128,64,77)$ at base and with subsequent planes having a spatial dimension multiplier of 2 and 4, respectively.  The $z$ dimension is halved compared to $x$ and $y$ because the rendering volume contains the cameras at the top of the volume pointing in approximately along the negative $z$ axis.  This saves on memory as the positive $z$ axis of any particular plane would remained unused.  The time dimension is set to half the number of total time steps in the scene.  Feature vector dimensions within each plane is set to 32.  
Additionally, we fine-tune the object detection model, YOLOv8 \cite{yolov8},  for 20 epochs with with a batch size of 16 and other settings follow the default settings in the official website \cite{yolov8}.

\subsection{Evaluation}

\subsubsection{Static NeRF}
To validate our method, we conducted experiments using Archangel-Real captured at diverse UAV altitudes. UAV-based images captured at different altitudes exhibit distinct image characteristics, presenting a challenging problem to tackle. Especially, at higher altitudes, the majority of human targets appear smaller, making it impractical to directly utilize detectors pre-trained by standard datasets. Object detectors trained for specific altitudes often struggle to generalize to images captured at different altitudes. Our data augmentation technique allows us to generate novel-view training images along with corresponding bounding boxes at different altitudes with unique viewpoints. We conducted four different experimental settings in which we initially trained the model with a low-altitude scenario (15m) and then evaluated it in high-altitude scenarios (35m, 40m, 45m, or 50m). In each experimental setup, we compare three detection models trained with different types of the dataset. ``Real" refers to the original data from Archangel-Real, while ``Synthetic" represents training data generated by NeRF. In terms of the synthetic data,  we generate the same number of training images as that of real data to ensure a fair comparison. The ``R+S" model utilizes a hybrid dataset, combining both ``Real" and ``Synthetic" datasets. Regarding object detection models, the YOLOv8 family \cite{yolov8} offers a range of five different levels of architectural complexity, all pretrained on MS-COCO \cite{MS-COCO}. We utilize YOLOv8 models with a lower number of model parameters, specifically YOLOv8n and YOLOv8s, suitable for operating on resource-constrained UAV platforms.

In Table \ref{table_static_nerf}, we can observe that YOLOv8 models trained with synthetic data only mostly outperform those trained solely with real datasets, indicating the advantage of synthesizing novel-view images at high altitudes and the realism of the synthetic data on par with the real data. Futhermore, when we optimize YOLOv8 on a hybrid dataset, it shows the best performance among models trained with either Real only or synthetic data only. Notably, the performance gap is even more noticeable at an altitude of 50m, thanks to the inclusion of high-fidelity synthetic images captured at high-altitudes. These results validate that data augmentation with NeRF can significantly enhance detection performance for UAV-based images, particularly in more challenging scenarios. In Table \ref{table_static_nerf2}, we initially trained the model at a high altitude of 50 meters and subsequently evaluated it in a low-altitude scenario at 15 meters. We observed a performance improvement trend similar to our previous results.

\begin{figure}[t]
\centering
\includegraphics[width=0.95\linewidth]{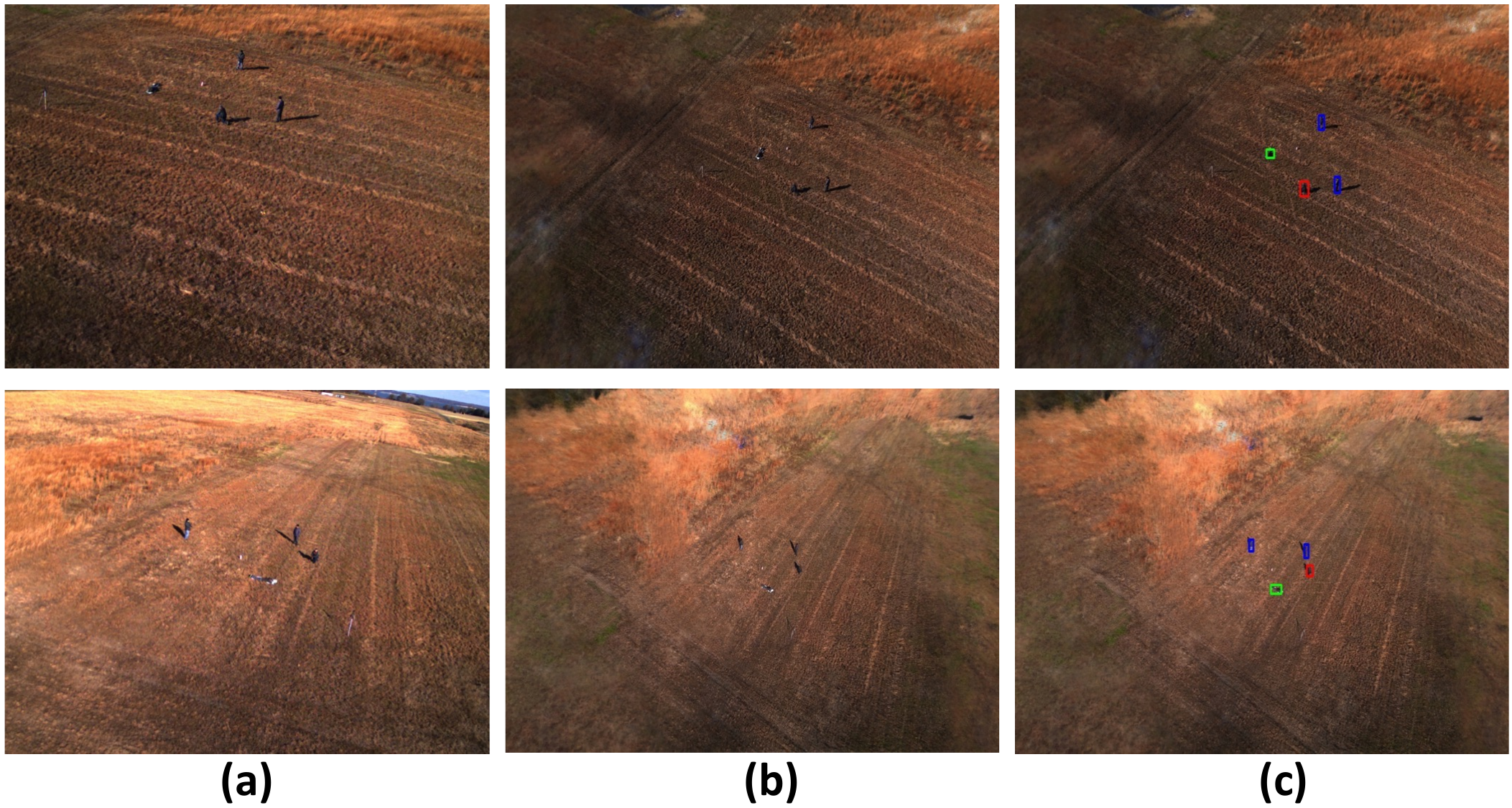}
\vspace{-3mm}
\caption{Visualization Results: (a) Examples of original images at 15m altitude from the Archangel dataset. (b) Synthetic images generated by NeRF. (c) Synthetic images with corresponding object pose: stand (blue), kneel (red), and prone (green). Although our NeRF model is initially trained on UAV images captured at low altitudes, it has the capability to generate photorealistic images from novel viewpoints at high altitudes.}
\label{Static_NeRF_figure3}
\vspace{-6mm}
\end{figure}

We present visualization results of our method in Figure \ref{Static_NeRF_figure3}. In Figure \ref{Static_NeRF_figure3}-(b), our approach demonstrates its ability to produce images with enhanced realism from novel viewpoints. Notably, we achieve this with only low-altitude UAV-based images (15m) used to train the NeRF model, enabling it to render realistic images captured at high altitudes. Additionally, as shown in Figure \ref{Static_NeRF_figure3}-(c), our pipeline can generate corresponding bounding boxes for these images.

\begin{table}[t]
    \begin{center}
    \caption{\label{table_static_nerf2}  AP Comparison on the different altitudes of YOLOv8n}
    \resizebox{0.45\textwidth}{!}{\normalsize
    \begin{tabular}{l|c|cccc|cccc}
    \toprule
    \multirow{2}{*}{Method} & \multirow{2}{*}{Altitude} & \multicolumn{4}{c|}{$\text{mAP}_{50}$} & \multicolumn{4}{c}{$\text{mAP}_{50:95}$} \\
    & &  Standing & Kneeling  & Prone & All & Standing & Kneeling & Prone & All \\ \hline
    Real & 50m\:\textrightarrow\:15m &  0.284 & 0.553 & 0.774 & 0.537 & 0.0623 & 0.122 & 0.263 & 0.149 \\
    Synthetic & 50m\:\textrightarrow\:15m & 0.432 & 0.684 & 0.898 & 0.671& 0.131 & 0.276 & 0.328 & 0.245 \\
    R + S & 50m\:\textrightarrow\:15m & \textbf{0.484} &  \textbf{0.727} & \textbf{0.929} & \textbf{0.713} & \textbf{0.152} & \textbf{0.275} & \textbf{0.362} & \textbf{0.263} \\
    \bottomrule
    \end{tabular}}
    \end{center}
    Our method shows improvement of YOLOv8n in $\text{mAP}_{50}$ metric by 32.77 \% and $\text{mAP}_{50:95}$ metric by 76.51 \% at 15m altitdue. 
    \vspace{-3mm}
\end{table}

\begin{figure}[t]
\centering
\vspace{-12mm}
\includegraphics[width=0.95\linewidth]{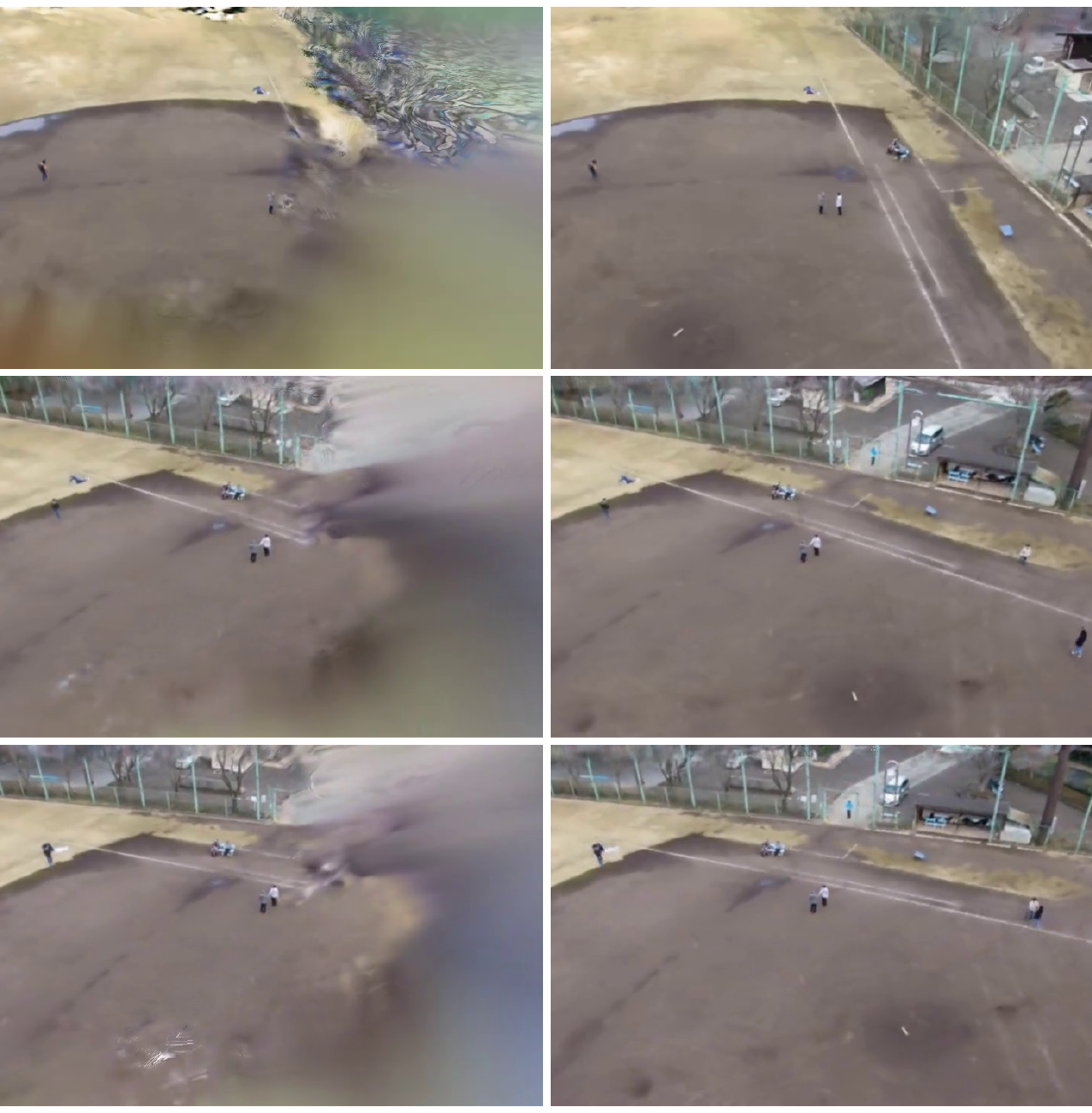}
\vspace{-13mm}
\caption{Left images show novel views from a validation set rendered by the stock K-Planes algorithm trained on a subset of the Okutama-Action dataset.  Right images show novel views from the same validation set but rendered with our extended K-Planes algorithm trained on the same subset of data and same hyperparameters.}
\label{KPlanes_NeRF_compare}
\vspace{-2mm}
\end{figure}

\begin{figure}[t]
\centering
\includegraphics[width=0.95\linewidth]{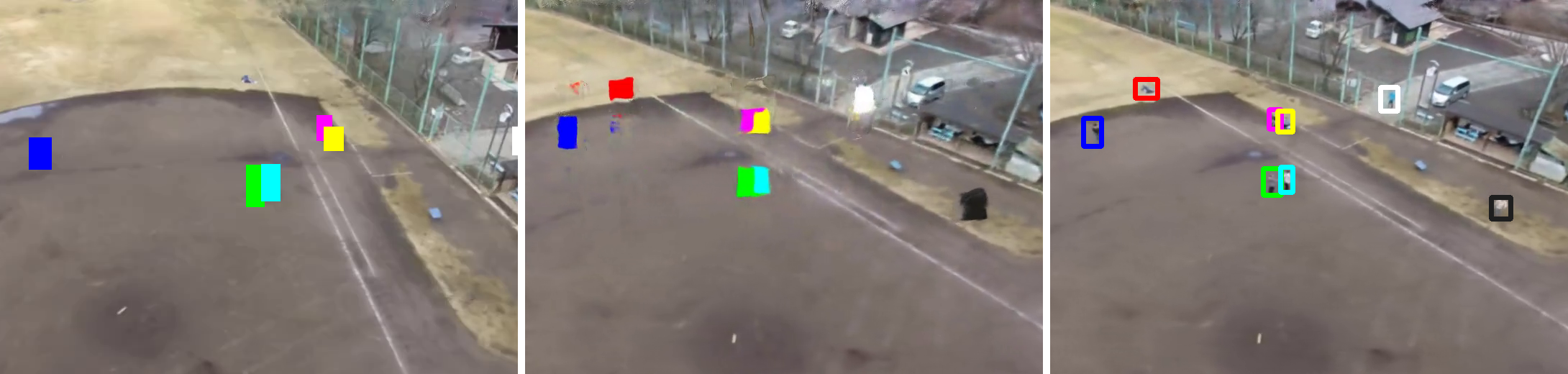}
\vspace{-1mm}
\caption{Visualization Results: (a) Example of masked image from the Okutama-Action dataset. (b) Novel-view images generated by extended K-Planes NeRF, with bounding box masks. (c) Novel-view images with corresponding bounding boxes: each person bound with a unique color box.}
\label{Dynamic_NeRF_figure}
\vspace{-6mm}
\end{figure}

\subsubsection{Dynamic NeRF}
For validation on dynamic scenes, we conducted experiments using a subset of Okutama-Action. Amongst the same challenges mentioned for Archangel, the irregular flight path and moving targets in Okutama make for an especially challenging dataset with respect to novel view synthesis.  Specifically, utilizing powerful techniques such as Pix-SfM \cite{pix-sfm} to extract camera poses can still output noisy pose estimates. Our extended version of K-Planes helps to alleviate some of these issues.


To further help minimize errors from pose estimates, we extract a subset of frames representing a smooth linear path from a single Okutama video, morning video from drone pilot 1, video 1.1.1, frames 230 to 535 for a total of 306 frames.  We then use Pix-SfM \cite{pix-sfm} to get the estimated poses for each frame and annotate each frame with a timestamp from 0 to 1 in their chronological order within the video.  These frames are separated into a train and validation set, taking every other frame for the training set.  Given this subset and training/validation split, we compare the PSNR of the validation set for both stock K-Planes and our extended K-Planes in Table \ref{table_dynamic_nerf1}.  Our K-Planes model increases PSNR by 3.85 points over stock K-Planes, a large improvement that can be seen qualitatively in Figure \ref{KPlanes_NeRF_compare}.  Our K-Planes model does a better job at separating the static and temporal elements of the scene given noisy camera pose estimations, reducing the blur and artifacts seen from stock K-Planes.

\begin{table}[t]
    \begin{center}
    \caption{\label{table_dynamic_nerf1} PSNR Comparison on Okutama-Action Validation Subset}
    \resizebox{0.2\textwidth}{!}{\normalsize
    \begin{tabular}{l|c}
    \toprule
    \multirow{1}{*}{Method} & \multirow{1}{*}{PSNR} \\
    \midrule
    Stock K-Planes & 16.62 \\
    Extended K-Planes & \textbf{20.47} \\ 
    \bottomrule
    \end{tabular}}
    \end{center}
    PSNR from validation subset of Okutama-Action.  Validation subset consistes of every other image from frames 230 to 535 from the video 1.1.1: morning video, drone pilot 1.
    \vspace{-3mm}
\end{table}

    

\begin{table}[t]
    \begin{center}
    \caption{\label{table_dynamic_nerf2} AP Comparison on the Okutama-Action of YOLOv8 family}
    \small
    \begin{tabular}{l|c|c|c}
        \toprule
        \multirow{1}{*}{Data} & \multirow{1}{*}{Model} & \multirow{1}{*}{$\text{mAP}_{50}$} & \multirow{1}{*}{$\text{mAP}_{50:95}$} \\
        \midrule
        Real & YOLOv8n & 0.234 & 0.072 \\
        Synthetic & YOLOv8n & 0.224 & 0.074 \\
        R + S & YOLOv8n & \textbf{0.263} & \textbf{0.075} \\
        \midrule
        Real & YOLOv8s & \textbf{0.256} & 0.086 \\
        Synthetic & YOLOv8s & \textbf{0.254} & \textbf{0.087} \\
        R + S & YOLOv8s & \textbf{0.255} & 0.082 \\
        \bottomrule
    \end{tabular}
    \end{center}
    ``Real" denotes the detection performance achieved through training with the subset of Okutama-Action data collected from video 1.1.1. ``Synthetic" represents the performance achieved through training with synthetic data generated by extended K-Planes NeRF. ``R+S" indicates the detection accuracy attained through training with a combined dataset of both ``real" and ``synthetic". Compared to YOLOv8n trained with "Real" dataset, we improve the $\text{mAP}_{50}$ metric by 12.4 \%.
    \vspace{-6mm}
\end{table}

The synthetic dynamic data is assessed in a similar manner to archangel by training on a subset of the Okutama-Action dataset and testing on equivalent test sets.  Three detection models trained with different dataset types, ``Real", ``Synthetic" and ``Real + Synthetic", are compared.  The test set for each model is the subset of test videos provided by Okutama that include drone pilot 1 footage during the morning, videos 1.1.8 and 1.1.9.

Table \ref{table_dynamic_nerf2} showcases the benefits we see with synthetic NeRF data when training YOLOv8 on a subset of Okutama-Action.  ``Synthetic" and ``real" trained models have equivalent performance, indicating that the ``synthetic" data rendered by our k-planes algorithm is effectively equivalent to the ``real" data from Okutama for training YOLOv8n/s.  For YOLOv8n, the ``R+S" model achieves a relative 12.4\% improvement in mAP50 over ``real" alone.  However, mAP50:95 does not improve significantly.  This may be due to errors in extracting bounding boxes from the NeRF model compared to manually labeled data.  We also do not see any improvements for YOLOv8s between any version of the models. This is likely due to general variation in training for YOLOv8 as well as the small extrapolations from training data for the dynamic ``synthetic" data.  The Archangel ``synthetic" data has larger variations in pose relative to its training data, so larger improvements are expected compared to the ``synthetic" data from Okutama-Action.  The ``synthetic" data from Okutama-Action is very similar to that of the ``real" training data so as to reduce errors and artifacts from extrapolating too far from the training data.  This is likely causing the larger YOLOv8s model, which has approximately three times as many parameters as that of YOLOv8n, to underfit during training.  From the perspective of UAV hardware and power limitations, the nano class models are more applicable and thus performance gains there are desirable over the other larger classes of models.

\section{Conclusions, Limitations, and Future Work} 

We have shown that NeRF algorithms are a valid method for bridging the domain gap, given a limited amount of real-world data, to supplement training for modern data-hungry algorithms, such as YOLOv8. We have also developed the model optimization pipeline where any SOTA detection model can be further optimized by unseen salient scene attributes captured with novel camera poses and self-generated bounding box annotation.  Given the current state of the art in NeRF, ideal UAV scenarios include circular paths with fixed camera angles, as shown by the performance improvements on the Archangel dataset as compared to Okutama-Action.  The challenges of UAV-based footage as well as dynamic scenes provide a limitation to the performance benefits with YOLOv8 on Okutama-Action; however, we do show that an increased PSNR can be achieved when compared to the stock K-Planes algorithm and a modest increase in mAP50 can be achieved for certain YOLOv8 models.  Future work may focus on fully exploring the extent of extrapolating ``synthetic" data from dynamic datasets.  The benefits shown in Archangel highlight the potential for improvements on Okutama-Action, while the increased PSNR from our extended K-Planes gives direction for future research in minimizing camera pose noise and better reconstructing dynamic UAV scenes, leading to enhanced recognition of an occurrence of certain activities or events of interest.


\textbf{Acknowledgements} 
This work was supported in part by the Office of the Under Secretary of Defense for Research and Engineering (OUSD R\&E) through Army Cooperative Agreement W911NF2120076, and ARO Grants W911NF2110026, W911NF2310046, and W911NF2310352.

\addtolength{\textheight}{0cm}   







\end{document}